\documentclass{article}

\usepackage{PRIMEarxiv}

\usepackage[utf8]{inputenc} % allow utf-8 input
\usepackage[T1]{fontenc}    % use 8-bit T1 fonts
\usepackage{hyperref}       % hyperlinks
\usepackage{url}            % simple URL typesetting
\usepackage{booktabs}       % professional-quality tables
\usepackage{amsfonts}       % blackboard math symbols
\usepackage{nicefrac}       % compact symbols for 1/2, etc.
\usepackage{microtype}      % microtypography
\usepackage{lipsum}
\usepackage{fancyhdr}       % header
\usepackage{graphicx}       % graphics
\graphicspath{{media/}}     % organize your images and other figures under media/ folder

\usepackage{cite}
\usepackage{amsmath,amssymb}
\usepackage{textcomp}
\usepackage{algorithm}%
\usepackage{algorithmicx}%
\usepackage{algpseudocode}%
\usepackage{listings}%
\usepackage{subcaption}

\title{TransFusion: Generating Long, High Fidelity Time Series using Diffusion Models with Transformers}

\author{
  Md Fahim Sikder, Resmi Ramachandranpillai, Fredrik Heintz \\
  Department of Computer and Information Science (IDA) \\
  Link\"oping University \\
  Link\"oping, Sweden\\
  \texttt{\{md.fahim.sikder, resmi.ramachandran.pillai, fredrik.heintz\}@liu.se} \\
}

\begin{document}
\maketitle

\begin{abstract}
The generation of high-quality, long-sequenced time-series data is essential due to its wide range of applications.  In the past, standalone Recurrent and Convolutional Neural Network-based Generative Adversarial Networks (GAN) were used to synthesize time-series data. However, they are inadequate for generating long sequences of time-series data due to limitations in the architecture. Furthermore, GANs are well known for their training instability and mode collapse problem. To address this, we propose TransFusion, a diffusion, and transformers-based generative model to generate high-quality long-sequence time-series data. We have stretched the sequence length to 384, and generated high-quality synthetic data. Also, we introduce two evaluation metrics to evaluate the quality of the synthetic data as well as its predictive characteristics. We evaluate TransFusion with a wide variety of visual and empirical metrics, and TransFusion outperforms the previous state-of-the-art by a significant margin.
\end{abstract}

% keywords can be removed
\keywords{Time Series Generation \and Generative Models \and Diffusion Models \and Synthetic Data \and Long-Sequenced Data}

%% main text
\section{Introduction}
\label{sec:introduction}

Synthetic time-series generation has been a popular research field in recent years due to its wide variety of applications. Electronic health record (EHR) generation, trajectory generation, and vehicle routing need tremendous amounts of data to learn the pattern in the samples. Unfortunately, it is challenging to get a hold of real data due to the lack of public access to the dataset \cite{walonoski2018synthea}. Synthetic data can be a solution to mitigate this problem \cite{torfi2020corgan}. However, generating synthetic time-series data is challenging due to its random nature. In recent years, researchers used different generative techniques like Variational Auto Encoders (VAE) and Generative Adversarial Networks (GAN) to generate synthetic time series data \cite{yoon2019time, niu2020lstm, xu2020cot, jeon2022gt}. But these studies were done using shorter sequence lengths (less than 100). Also, training GAN is extremely unstable, and it is prone to the mode-collapse problem \cite{goodfellow2020generative}, where the model provides limited sample variety. Generating long-sequenced time-series data is necessary to understand the data context better. The longer the sequence length is, the more information it can capture about the data over time. For example, monitoring a diabetic patient's blood sugar level for a long time will reveal patterns and trends better than monitoring for a short period.

Capturing temporal dependencies in long-sequential data is challenging for standalone Convolutional Neural Networks (CNNs) and Recurrent Neural Networks (RNNs) because of their architectural limitations \cite{ismail2019deep, siami2019performance}. These architectures fail to capture the long-term dependencies because they can not look back in sequences for datasets with longer sequence lengths. Also, these architectures possess some computational challenges, for example, in the case of RNNs, parallelization is not possible due to sequential computational dependencies \cite{rasul2021autoregressive}. Moreover, most of the time series generative models are GAN-based and are prone to mode-collapse problems, which limits their ability to produce high-fidelity synthetic data \cite{goodfellow2020generative}.

%Most of the evaluation metrics for synthetic time series are also CNN and RNN-based. So, new metrics are needed for evaluating synthetic time series with long sequences.

Diffusion-based models have garnered extensive interest for their potential in the Computer Vision (CV) and Natural Language Processing (NLP) domains \cite{ho2020denoising, nichol2021improved}. 
%These models gradually add noise to the input data and make it a noisy data distribution. 
The whole process involves adding noise to the data, called the forward diffusion process, and a neural network architecture which is trained to learn the data distribution of the real data from the noisy distribution. This latter is called the backward diffusion process. After training, the diffusion-based generative model can generate high-quality synthetic data. Additionally, these models overcome the mode-collapse problem by learning the semantic nature of the data. Though there have been works in the computer vision and NLP domains using diffusion models, time-series generation is still unexplored in the context of high dimensionality, long sequence length, and diffusion models. 

In this study, we present TransFusion, a diffusion-based model to generate high-quality time series data. We use a transformers encoder \cite{vaswani2017attention, liu2022memory} in the backward diffusion process to approximate the real data distribution. As the transformer encoder consists of the attention mechanism, it can capture the long-term temporal dependencies. We conduct extensive experiments with the data consisting of sequences of length 100. We further investigate the generated data quality by TransFusion on the sequence length 384. We perform various evaluations using visual (PCA \& t-SNE) and empirical metrics. Moreover, most state-of-the-art metrics rely on CNN or RNN architectures, limiting their effectiveness in capturing and measuring long sequences. This underscores the significance of new evaluation metrics designed explicitly for evaluating synthetic time series with extended sequences. In contrast, we propose two evaluation metrics based on transformers, enabling them to capture high-dimensional relationships. These metrics distinguish between synthetic and original data (post-hoc classifier) and see if the synthetic data preserves the predictive characteristics of the original data by using a sequence prediction task. After extensive experiments, we find that TransFusion outperforms the state-of-the-art methods in these evaluations and generates high-fidelity, long-sequence time-series data.

%Since existing evaluation metrics for synthetic data are based on RNN and CNN architectures, they are unable to capture long-term dependencies. Therefore, we propose two transformers-based evaluation metrics to

The main contributions of our proposed approach are the following:

\begin{itemize}
	\item We introduce TransFusion, a diffusion-based generative model, capable of generating long-sequence, high-fidelity time series data exploiting the power of transformers that outperforms previous state-of-the-art time series generative models, such as GT-GAN \cite{jeon2022gt}, TimeGAN \cite{yoon2019time}, CotGAN \cite{xu2020cot}, TTS-GAN \cite{li2022tts}, TimeGrad \cite{rasul2021autoregressive}, etc.
	\item We propose two evaluation metrics for synthetic time series data which can distinguish original and synthetic data and provide an overview of the synthetic data's performance over sequence prediction tasks. 
	\item We show the quality of the generated data using a wide variety of evaluation metrics (both visual and empirical).
	\item We conduct an ablation study to empirically demonstrate that the combined architecture (diffusion+transformers) is needed for generating high-quality, long-sequenced time series data.
\end{itemize}

The code and supplementary materials can be found at \url{https://github.com/fahim-sikder/TransFusion}.

\section{Related Works}
\label{sec:relatedworks}

Time-series generation has seen increased attention in recent years for its diverse applications. In most state-of-the-art works, Generative Adversarial Networks (GAN) have been used to generate synthetic time series data \cite{yoon2019time, donahueadversarial}. GAN has been widely used in computer vision (CV) and Natural Language Processing (NLP) domains with a high success rate \cite{bao2017cvae, zhang2017adversarial}. For the time series generation, most of the studies used Recurrent Neural Network (RNN) and Convolutional Neural Network (CNN) based architecture. For example, TimeGAN \cite{yoon2019time}, used a combination of supervised and unsupervised learning to generate time series data, and in the architecture, they used Recurrent Neural Networks. Causal optimal transport was used to generate the time sequence in CotGAN \cite{xu2020cot}. WaveGAN \cite{donahueadversarial} used a combination of Transposed Convolutional Neural Networks in its generator for generating audio data and QuantGAN \cite{wiese2020quant} used Temporal Convolutional Networks (TCN) \cite{lea2017temporal} to generate financial time-series data. Sig-WGAN \cite{ni2021sig} is a time-series generative model, that also generates financial time-series data combining a continuous-time probabilistic model with W1 metric. TTS-GAN \cite{li2022tts} used transformers encoder combined with GANs to generate time-series data. GT-GAN \cite{jeon2022gt} used a combination of GANs, Autoencoders, Neural Ordinary Differential Equations, Neural Controlled Differential Equations, and Continuous time-flow processes for time series generation. Although standalone RNN and CNN architectures have been used widely, these architectures are not well suited for generating longer sequences, because they cannot capture the long-term time dependencies. So, all of the studies mentioned above cannot generate meaningful time series data when the sequence length is longer, ie., a sequence length of more than 100 (we show in Table \ref{tab:evaluation-table1} and \ref{tab:evaluation-table2}). Apart from all the aforementioned problems, GANs are prone to the mode collapse problem, which is critical in time series domains.

In contrast, the diffusion models have recently yielded promising results in both the CV and NLP domains. Diffusion-based generative models are capable of generating high-quality samples. These models are not prone to mode collapse which provides an advantage over the GAN-based models. The generation of time series using diffusion-based architectures is a fairly new idea. It has been tested on conditional and unconditional video generation \cite{ho2022video} tasks, and audio generation tasks \cite{kong2020diffwave}. TimeGrad \cite{rasul2021autoregressive} used autoregressive DDPM to forecast multivariate probabilistic time-series. In their model, they used RNN-based architecture to capture the temporal dependencies. In Diffwave \cite{kong2020diffwave}, the author used a Convolutional Neural Network-based diffusion model that can synthesize audio and outperforms previous GAN-based audio synthesis work. However, it is important to note that most audio data is low dimensional. This is in contrast to general time-series data, which can often be high-dimensional.

In this study, we leverage the use of transformer architecture to capture the long-term dependencies of time and combine them with the diffusion model. Also, to distinguish original data from synthetic data for long sequences, we propose a transformer-based post-hoc classifier. Additionally, to analyze the predictive characteristics of the synthetic data, we propose a metric called long-sequence predictive score (LPS) by exploiting the transformer's ability to capture long relationships.

\section{Preliminaries}
\label{sec:preliminaries}

In this section, we formulate the problem definition and give the necessary background information for the proposed approach.

\subsection{Problem Statement}
\label{subsec:problemstatement}

Given a time series data $x \sim q(x)\in \mathbb{R}^D$, where $x = \{x^1, x^2, x^3, ..., x^N\}$, $N$ is the sequence length of time series data, and $D$ is the data dimension, the objective is to train a generative model that will learn $p_\theta(x)$ which approximate $q(x)$. 

%Our proposed approach uses diffusion-based models for synthesizing time-series data.

\subsection{Denoising Diffusion Probabilistic Model (DDPM)}

The denoising diffusion probabilistic model \cite{ho2020denoising} works in two steps. First, it takes data $x_0 \sim q(x_0)$ and adds Gaussian noise with a timestamp of $T$ steps, to obtain the Gaussian noise $x_T$ (if $T$ is higher \cite{nichol2021improved}). The amount of noise added to each step is calculated by the variance schedule $\beta_{t}$. So the forward diffusion process can be defined as follows:

\begin{equation}
	q(x_{1:T}|x_0) = \prod_{t = 1}^T q(x_t|x_{t-1})
\end{equation}

\begin{equation}
	q(x_t|x_{t-1}) = \mathcal{N}(x_t;\sqrt{1-\beta_t}x_{t-1}, \beta_t\textbf{I})
\end{equation}
\begin{equation}
	q(x_t|x_0) = \mathcal{N}(x_t; \sqrt{\Bar{\alpha_t}}x_0, (1-\Bar{\alpha_t})\textbf{I})
\end{equation}

Here, $\alpha_t = 1 - \beta_t$ and $\Bar{\alpha_t} = \prod_{s=0}^t \alpha_s$. In the backward diffusion process, a model is trained to learn $p_\theta$ that approximate $q(x_{t-1}|x_t)$ as it is difficult to learn $q(x_{t-1}|x_t)$ directly as it needs the knowledge of the real data distribution.

\begin{equation}
	\label{eq:diff-back}
	p_\theta(x_{t-1}|x_t) = \mathcal{N}(x_{t-1}, \mu_\theta(x_t, t), \Sigma_\theta(x_t, t))
\end{equation}

In practice, a neural network is employed to learn the distribution $p_\theta(x_{t-1}|x_t)$, and it tries to predict the parameters $\mu_\theta(x_t, t)$ and $\Sigma_\theta(x_t, t)$. DDPM \cite{ho2020denoising} achieved the best result with a fixed variance $\sigma^2\textbf{I}$. And the mean $\mu_\theta(x_t, t)$ is calculated as follows:

\begin{equation}
	\mu_\theta(x_t, t) = \frac{1}{\sqrt{\alpha_t}}\big(x_t-\frac{\beta_t}{\sqrt{1-\Bar{\alpha_t}}}\epsilon_{\theta}(x_t, t)\big)
\end{equation}

So we can re-write equation \ref{eq:diff-back} as:

\begin{equation}
	\label{eq:approx}
	\begin{split}
		p_\theta(x_{t-1}|x_t) = \mathcal{N}(x_{t-1},\frac{1}{\sqrt{\alpha_t}}\big(x_t-\frac{\beta_t}{\sqrt{1-\Bar{\alpha_t}}} & \epsilon_{\theta}(x_t, t)\big), \\
		& \sigma^2\textbf{I})
	\end{split}
\end{equation}

Also the training objective defined by DDPM is,

\begin{equation}
	L_{simple} = \mathbb{E}_{t,x_0,\epsilon}\big[||\epsilon - \epsilon_\theta(x_t, t)||^2\big]
	\label{eq:obje}
\end{equation}

where, $\epsilon \sim \mathcal{N}(0, \textbf{I})$

\section{TransFusion: Transformers Diffusion Model}
\label{sec:model}

\begin{figure}[!t]
	\centering
	\includegraphics[width = 0.6\textwidth]{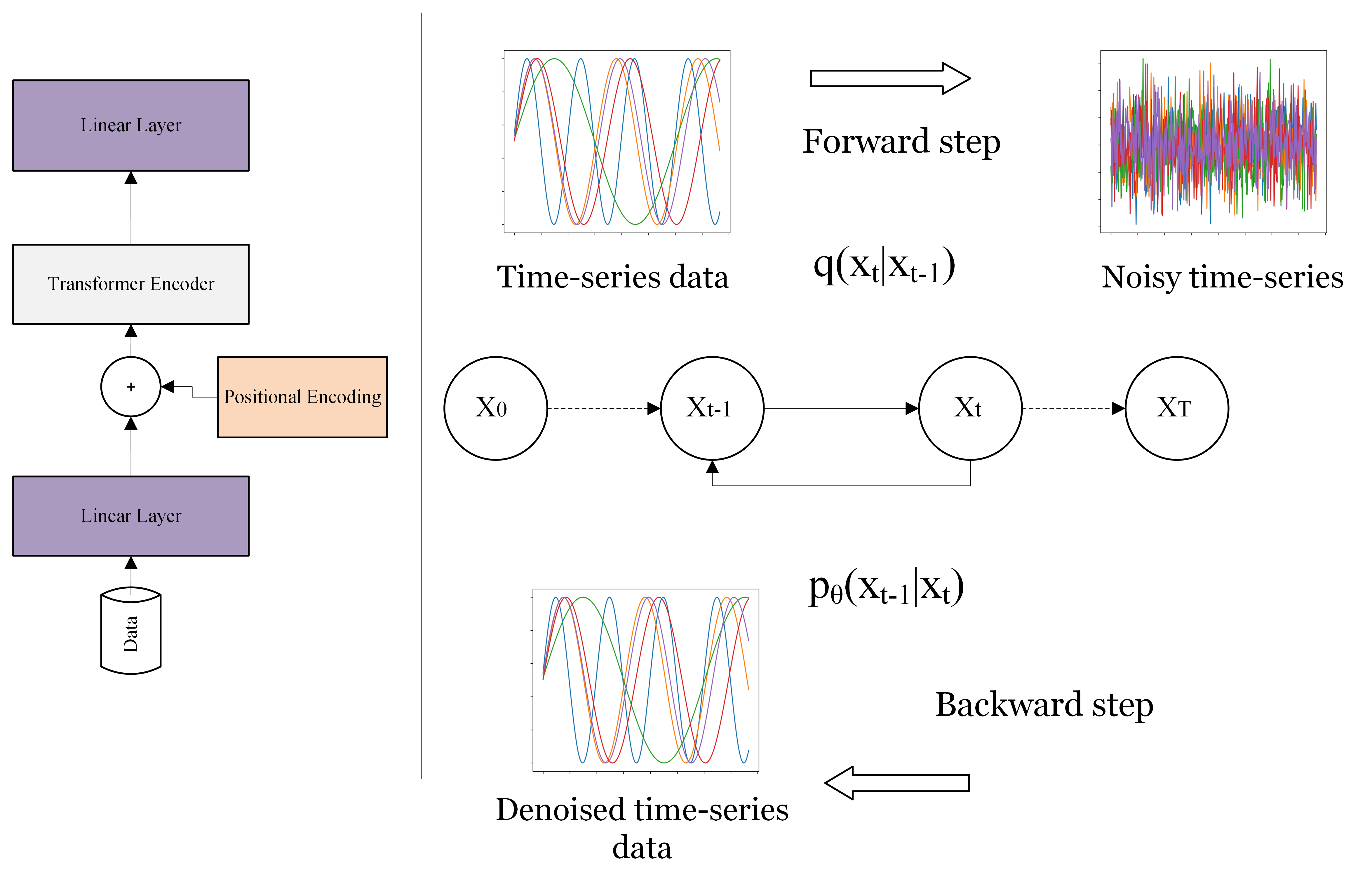}
	\caption{TransFusion Architecture, Transformers block that used in the TransFusion (left part), working process of diffusion (right part)}
	\label{fig:transfusion}
\end{figure}

In this study, we introduce a diffusion-based time-series generative model called TransFusion. We use the transformer encoder as the neural network architecture to approximate the target data distribution for the backward diffusion process. Figure \ref{fig:transfusion} shows the overall architecture of the TransFusion. 

\subsection{Training and Inference}
\label{sec:training}

For the forward diffusion step, we use the cosine variance scheduler \cite{nichol2021improved} to calculate the amount of noise added in each time-step. Originally, these variance schedulers (Linear, Cosine, Sigmoid, etc.) were designed to work with images, but as the ultimate goal of these schedulers is to inject noise gradually until the data becomes pure noise so, it can also be used for the time-series data \cite{rasul2021autoregressive, ho2020denoising}.

After adding noise to the data, in the backward diffusion steps, we take the neural network architecture to denoise the data. The neural network architecture takes the noisy time-series data that first goes through a linear layer follows by the positional encoding, then to the transformer encoder. Positional encoding is used to keep track of the data in each time-step. Then the data is passed through another linear layer to make it the same shape as the original data. 

Over time, the network tries to denoise the data and learns to approximate the original data distribution using the objective equation \ref{eq:obje}. Algorithm \ref{alg:train} \cite{ho2020denoising} show how TransFusion is trained. For the inference, we first sample noise from the Gaussian distribution and then use the trained neural network to generate synthetic data samples using the noise. Algorithm \ref{alg:sample} shows inference steps.

\subsection{Implementation Setup}

We use PyTorch to implement TransFusion. For the architecture, we take a hidden dimension of $256$, a batch size of $256$, attention head is $8$ and $6$  transformer encoder layers. We use the $Adam$ optimizer with a learning rate of $1e-4$. We train TransFusion for $5000$ epochs on each datasets.

For the evaluation, we use held-out dataset and compare it with the synthetic data generated by the models.

\begin{algorithm}
	\caption{Training}
	\label{alg:train}
	\begin{algorithmic}
		\State {\bfseries Input:} Pre-calculated variance schedule $\beta$
		\Repeat
		\State sample $x_0 \sim q(x_0)$, $\epsilon \sim \mathcal{N}(0, \textbf{I})$, $t \sim \mathcal{U}(\{1,..,T\})$
		\State Take Gradient step on
		\State $\nabla_\theta||\epsilon - \epsilon_\theta(x_t, t)||^2$
		\Until{converged}
	\end{algorithmic}
\end{algorithm}

\begin{algorithm}
	\caption{Sampling}
	\label{alg:sample}
	\begin{algorithmic}
		% \STATE {\bfseries Input:} Pre-calculated variance schedule $\beta$
		\State sample $x_T \sim \mathcal{N}(0, \textbf{I})$
		\For{$t=T, ..., 1$}
		\State sample $x_{t-1} \sim p_\theta(x_{t-1|x_t})$ using equation \ref{eq:approx}
		\EndFor
		\State {\bfseries return:} $x_0$
	\end{algorithmic}
\end{algorithm}

\section{Experimental Results}
\label{sec:experiment}

%With four time-series datasets, we compare our results with five time-series generative models, TimeGAN \cite{yoon2019time}, CotGAN \cite{xu2020cot}, WaveGAN \cite{donahueadversarial}, QuantGAN \cite{wiese2020quant}, and GT-GAN \cite{jeon2022gt}. The experiments use sequence lengths of $100$ and $384$. It is shown that TransFusion is capable of producing high-quality synthetic data even at such a long sequence length (384). Qualitative and quantitative evaluation metrics are used to evaluate our study. In addition, we propose two metrics for assessing the predictive nature of the synthetic data and comparing it with the original data. The combination of diffusion and transformers is also necessary to generate long-sequenced time-series data, as shown in an ablation analysis. %Datasets preprocessing procedure, hyperparameters of the TransFusion and benchmarking models can be found in the appendices \ref{ap:dataset}, \ref{ap:hyper}.

We perform extensive experiments with four time-series datasets and compare the results with eight time-series generative models, TimeGAN \cite{yoon2019time}, CotGAN \cite{xu2020cot}, WaveGAN \cite{donahueadversarial}, QuantGAN \cite{wiese2020quant}, Sig-WGAN \cite{ni2021sig}, TTS-GAN \cite{li2022tts}, TimeGrad \cite{rasul2021autoregressive}, and GT-GAN \cite{jeon2022gt}. We use publicly available source code to implement the benchmarks. We use sequence lengths $100$ and $384$ for the experiments. The result shows that even at this high sequence length (384), TransFusion is capable of generating high-quality synthetic data. We evaluate our study with various evaluation metrics, including qualitative and quantitative analysis. We also propose two evaluation metrics to assess the predictive nature of the synthetic data and to evaluate if the original data and the synthetic data are distinguishable. We also present an ablation analysis to show that the combination of diffusion and transformers is necessary to generate long-sequenced time-series data.

\begin{table*}[]
	\caption{Resutls on Different Time-Series Datasets, Sequence Length: 100, we report the mean and standard deviation (bold indicates best result, $\uparrow$ indicates higher as better, $\downarrow$ indicates lower as better)}
	\centering
	\resizebox{1.0\columnwidth}{!}{%
		\begin{tabular}{lcccccccc}\toprule
			& \multicolumn{7}{c}{\textbf{Evaluation Metrics}} 
			\\\cmidrule(lr){3-9}
			\textbf{Dataset} & \textbf{Model Name} & LDS & LPS & JSD & $\alpha$- & $\beta$- & Coverage & + 5 Steps\\ 
			& &  & & & precision & recall & & Ahead\\ 
			& & ($\downarrow$) & ($\downarrow$) & ($\downarrow$) & ($\uparrow$)& ($\uparrow$)& ($\uparrow$) & ($\downarrow$)\\ \midrule
			& TimeGAN & 0.499  $\pm$ .001 & 0.253 $\pm$ .001 & 0.11 $\pm$ .00 &  0.863 $\pm$ .00 & 0.089 $\pm$ .00 & 0.355 $\pm$ .00 & 0.257 $\pm$ .005\\ 
			& CotGAN &  0.467 $\pm$ .003  & 0.249 $\pm$ .001 & 0.10 $\pm$ .00 &  0.795 $\pm$ .00 & \textbf{0.984} $\pm$ .00 & 0.852 $\pm$ .00 & 0.253 $\pm$ .001\\ 
			%			& C-RNN-GAN  &  0.5 $\pm$ .00  & 0.649  $\pm$ .001 & 0.53 $\pm$ .00 & 0.00 $\pm$ .00 & 0.00 $\pm$ .00 & 0.00 $\pm$ .00 & 0.651 $\pm$ .001 \\
			%			Sine & EBGAN & 0.499 $\pm$ .001  &0.260 $\pm$ .001 & 0.32 $\pm$ .00 & 0.901 $\pm$ .00 & 0.00 $\pm$ .00 & 0.322 $\pm$ .00 & 0.261 $\pm$ .001\\ 
			& WaveGAN  & 0.500 $\pm$ .000 & 0.353 $\pm$ .001& 0.414 $\pm$ .00 & 0.00 $\pm$ .00 & 0.00 $\pm$ .00 & 0.000 $\pm$ .00 & 0.351 $\pm$ .001\\
			& QuantGAN  & 0.469 $\pm$ .013 & 0.249 $\pm$ .001 & 0.253 $\pm$ .00 & \textbf{0.992} $\pm$ .00 & 0.384 $\pm$ .00 & 0.823 $\pm$ .00 & 0.253 $\pm$ .001\\
			Sine    & Sig-WGAN  & 0.490 $\pm$ .001 & 0.249 $\pm$ .001 & 0.567 $\pm$ .00 & 0.944 $\pm$ .00 & 0.199 $\pm$ .00 & 0.833 $\pm$ .00 & 0.254 $\pm$ .001\\
			& TTS-GAN  &  0.5 $\pm$ .00 & 0.578 $\pm$ .086 & 0.407 $\pm$ .00 &  0.00 $\pm$ .00 & 0.00 $\pm$ .00 & 0.00 $\pm$ .00 &  0.528 $\pm$ .053\\
			& TimeGrad  &  0.489 $\pm$ .00 & 0.361 $\pm$ .001 & 0.311  $\pm$ .00 & 0.00 $\pm$ .00 & 0.00 $\pm$ .00 &  0.00 $\pm$ .00 & 0.365 $\pm$ .001\\
			& GT-GAN  & 0.465  $\pm$ .001 & 0.246 $\pm$ .001 & 0.048 $\pm$ .00 & 0.684 $\pm$ .00 &  0.954 $\pm$ .00 & 0.737 $\pm$ .00 & 0.253 $\pm$ .001 \\
			& TransFusion (ours)  & \textbf{0.097} $\pm$ .042 & \textbf{0.243} $\pm$ .001 & \textbf{0.02} $\pm$ .00 & \textbf{0.992} $\pm$ .00 &  0.974 $\pm$ .00 & \textbf{0.963} $\pm$ .00 & \textbf{0.245} $\pm$ .001\\ \midrule
			& TimeGAN  &  0.494 $\pm$ .001 & 0.062 $\pm$ .001 & 0.11 $\pm$ .00 & 0.994 $\pm$ .00 & 0.649 $\pm$ .00 & 0.073 $\pm$ .00 & 0.063 $\pm$ .001\\ 
			& CotGAN &   0.494 $\pm$ .001 &0.066 $\pm$ .001  & \textbf{0.010} $\pm$ .00 &  0.827 $\pm$ .00 & 0.837 $\pm$ .00 & 0.295 $\pm$ .00 & 0.067 $\pm$ .001\\ 
			%			& C-RNN-GAN & \textbf{0.19}  $\pm$ .03 & 0.068 $\pm$ .001 & 0.05 $\pm$ .00 & 0.899 $\pm$ .00 & 0.963 $\pm$ .00 & 0.605 $\pm$ .00 & 0.069 $\pm$ .001 \\ 
			%			Stock & EBGAN & 0.5  $\pm$ .00  & 0.495 $\pm$ .001& 0.66 $\pm$ .00 & 0.487 $\pm$ .00 & 0.00 $\pm$ .00 & 0.011 $\pm$ .00 & 0.496 $\pm$ .001 \\
			& WaveGAN  & 0.496 $\pm$ .007 & 0.064 $\pm$ .000& 0.272 $\pm$ .00 & \textbf{1.00} $\pm$ .00 & 0.000 $\pm$ .00 & 0.010 $\pm$ .00 & 0.064 $\pm$ .001\\
			& QuantGAN  & 0.497 $\pm$ .003 & 0.062 $\pm$ .000& 0.824 $\pm$ .00 & 0.660 $\pm$ .00 & 0.826 $\pm$ .00 & 0.405 $\pm$ .00 & 0.063 $\pm$ .001 \\
			Stock			& Sig-WGAN  & 0.5 $\pm$ .00 & 0.105 $\pm$ .05 & 0.797 $\pm$ .00 & 0.00 $\pm$ .00 & \textbf{1.000} $\pm$ .00 & 0.00 $\pm$ .00 & 0.103 $\pm$ .02\\
			& TTS-GAN  & 0.482 $\pm$ .00 & 0.067 $\pm$ .001 & 0.811 $\pm$ .00 & \textbf{1.00} $\pm$ .00 & 0.035 $\pm$ .00 & 0.067 $\pm$ .00 &  0.068 $\pm$ .001\\
			& TimeGrad  &  0.5 $\pm$ .00 &0.886 $\pm$ .001 &0.147  $\pm$ .00 &  0.00 $\pm$ .00 &  0.00 $\pm$ .00 &  0.00 $\pm$ .00 &0.847  $\pm$ .09\\
			& GT-GAN  & \textbf{0.416}  $\pm$ .026 & 0.071 $\pm$ .001 & 0.048 $\pm$ .00 & 0.684 $\pm$ .00 &  0.954 $\pm$ .00 & 0.737 $\pm$ .00 & 0.071$\pm$ .001 \\
			& TransFusion (ours)  & 0.498 $\pm$ .001 &\textbf{0.060} $\pm$ .001 & 0.018 $\pm$ .00 & 0.97 $\pm$ .00 &  0.99 $\pm$ .00  &\textbf{ 0.93} $\pm$ .00 & \textbf{0.061} $\pm$ .001\\ \midrule
			~ & TimeGAN &  0.497 $\pm$ .001 & 0.002 $\pm$ .001 & 0.11 $\pm$ .00 &  0.916 $\pm$ .00 &  0.343 $\pm$ .00 & 0.278 $\pm$ .00 & 0.003 $\pm$ .001\\ 
			& CotGAN & 0.482 $\pm$  .017 & 0.006 $\pm$ .001 &  0.10 $\pm$ .00 &  0.850 $\pm$ .00 & 0.617 $\pm$ .00 & 0.687 $\pm$ .00 & 0.006 $\pm$ .001 \\ 
			%			& C-RNN-GAN &  0.5 $\pm$  0.00 & 0.061 $\pm$ .001 & 0.344 $\pm$ .00 & \textbf{1.00} $\pm$ .00 & 0.00 $\pm$ .00 & 0.001 $\pm$ .00 & 0.065 $\pm$ .001\\ 
			%			Air & EBGAN &  0.5 $\pm$ .00 & 0.872 $\pm$  .106 & 0.423 $\pm$ .00 & 0.00 $\pm$ .00 & 0.00 $\pm$ .001 & 0.00 $\pm$ .00 & 0.813 $\pm$ .001\\
			& WaveGAN  & 0.439 $\pm$ .034 & 0.009 $\pm$ .001& 0.595 $\pm$ .00 & 0.942 $\pm$ .00 & 0.586 $\pm$ .00 & 0.750 $\pm$ .00 & 0.010 $\pm$ .001 \\
			& QuantGAN  & 0.464 $\pm$ .030 & 0.002 $\pm$ .001& 0.555 $\pm$ .00 & \textbf{0.983} $\pm$ .00 & 0.506 $\pm$ .00 & 0.749 $\pm$ .00 & \textbf{0.002} $\pm$ .001\\
			Air 			& Sig-WGAN  & 0.5 $\pm$ .00 & 0.064 $\pm$ .001 & 0.512 $\pm$ .00 & 0.261 $\pm$ .00 & 0.002 $\pm$ .00 & 0.001 $\pm$ .00 & 0.064 $\pm$ .001\\
			& TTS-GAN  &  0.5 $\pm$ .00 & 0.934 $\pm$ .001 &  0.528$\pm$ .00 &  0.00 $\pm$ .00 & 0.00 $\pm$ .00 &  0.00 $\pm$ .00 & 0.934 $\pm$ .001\\
			& TimeGrad  & 0.498 $\pm$ .001 & 0.359 $\pm$ .202 & 0.831 $\pm$ .00 &  0.00 $\pm$ .00 &  0.018 $\pm$ .00 &  0.00 $\pm$ .00 & 0.554 $\pm$ .211\\
			& GT-GAN  &  0.491 $\pm$ .005 & 0.003 $\pm$ .001 & 0.065  $\pm$ .00 & 0.687 $\pm$ .00 &  0.938 $\pm$ .00 & 0.824 $\pm$ .00 & 0.003 $\pm$ .001\\
			& TransFusion (ours)  & \textbf{0.148} $\pm$ .029 & \textbf{0.001} $\pm$ .001& \textbf{0.027} $\pm$ .00 & 0.926 $\pm$ .00 & \textbf{0.956} $\pm$ .00 & \textbf{0.944} $\pm$ .00 & \textbf{0.002} $\pm$ .001\\ \midrule
			~ & TimeGAN  &  0.490 $\pm$ .001 & 0.015 $\pm$ .001 & 0.176 $\pm$ .00 &  0.816 $\pm$ .00 &  0.213 $\pm$ .00 & 0.076 $\pm$ .00 & 0.015 $\pm$ .001\\ 
			& CotGAN &  0.497 $\pm$ .002 & 0.066 $\pm$ .001 & 0.183 $\pm$ .00 &  0.968 $\pm$ .00 & 0.374 $\pm$ .00 & 0.28 $\pm$ .00 & 0.067 $\pm$ .001\\ 
			%			& C-RNN-GAN  & 0.5 $\pm$ .00 & 0.264 $\pm$ .001 & 0.170 $\pm$ .34 & 0.00 $\pm$ .00 &  0.00 $\pm$ .00 & 0.00 $\pm$ .00 & 0.215 $\pm$ .001 \\ 
			%			Energy & EBGAN  & 0.5 $\pm$ 0.00 &0.971 $\pm$ .001 & 0.49 $\pm$ .00 &  0.00 $\pm$ .00 &  0.00 $\pm$ .00 & 0.00 $\pm$ .00 & 0.971 $\pm$ .001 \\
			& WaveGAN  & 0.496 $\pm$ .010 & 0.012 $\pm$ .000& 0.211 $\pm$ .00 & \textbf{1.00} $\pm$ .00 & 0.00 $\pm$ .00 & 0.014 $\pm$ .00 & \textbf{0.012 }$\pm$ .001 \\
			& QuantGAN  & 0.498 $\pm$ .002 & 0.012 $\pm$ .000& 0.826 $\pm$ .00 & 0.793 $\pm$ .00 & 0.701 $\pm$ .00 & 0.373 $\pm$ .00 & \textbf{0.012} $\pm$ .001 \\
			Energy			& Sig-WGAN  & 0.5 $\pm$ .00 & 0.027 $\pm$ .001 & 0.798 $\pm$ .00 & 0.00 $\pm$ .00 & 0.00 $\pm$ .00 & 0.00 $\pm$ .00 & 0.028 $\pm$ .001\\
			& TTS-GAN  & 0.5  $\pm$ .00 & 0.389 $\pm$ .237 & 0.796 $\pm$ .00 &  0.00 $\pm$ .00 & 0.00 $\pm$ .00 & 0.00 $\pm$ .00 &  0.437 $\pm$ .299\\
			& TimeGrad  &  0.5 $\pm$ .00 & 0.971 $\pm$ .001 & 0.152 $\pm$ .00 &  0.00 $\pm$ .00 &  0.00 $\pm$ .00 &  0.00 $\pm$ .00 &  0.971 $\pm$ .001\\			
			& GT-GAN  & 0.423 $\pm$ .030 & 0.012 $\pm$ .001& 0.158 $\pm$ .00 & 0.820 $\pm$ .00 & 0.26 $\pm$ .00 & 0.391 $\pm$ .00 & \textbf{0.012} $\pm$ .001 \\
			& TransFusion (ours)  & \textbf{0.332}  $\pm$ .019 & \textbf{0.011} $\pm$ .001 & \textbf{0.016} $\pm$ .00 & 0.983 $\pm$ .00 & \textbf{0.876} $\pm$ .00 & \textbf{0.847} $\pm$ .00 & \textbf{0.012} $\pm$ .001\\ \bottomrule
		\end{tabular}
	}
	\label{tab:evaluation-table1}
	
\end{table*}

\subsection{Datasets}

We use four time-series datasets for this experiment (one simulated and three real-world datasets).

\subsubsection{Sine Data}

We create a sinusoidal function $f(x)$ that generates sine wave data with a dimension of five with various frequencies and phases. The function generates $x_i(t) = sin(2\pi ft + \phi)$, here $f$ represents the frequency and it is sampled from a uniform distribution, $f\sim \mathcal{U}[0, 1]$, the phase $\phi$ is as well sampled from the uniform distributions, $\phi \sim \mathcal{U}[-\pi, \pi]$. The phase is different in each dimension $i \in {1,2, ..., 5}$.

\subsubsection{Stock}

We use Google stock data for the time between 2004 to 2017. This dataset contains six stock features (opening, closing, highest, lowest, and adjacent closing values) per day.

\subsubsection{Energy Consumption}

We also test our model using high-dimensional, noisy energy consumption data from the UCI repository \cite{candanedo2017data}. This dataset contains the electricity consumption of households at 10 minutes intervals for 4.5 months. It has 28 features, including electricity consumption in kilowatts per hour, temperature, humidity reading in various rooms, wind speed, pressure, etc.

\subsubsection{Air Quality}

Finally, we consider using another high-dimensional dataset on air quality \cite{de2008field} from the UCI repository. This dataset was collected from an Italian city hourly and recorded for one year. The data contains 13 features, including the level of carbon mono-oxide (CO), benzin level, etc. in the air.

\subsection{Evaluation Metrics}

%Synthetic data quality is measured by fidelity (does it come from the same distributions as the original one), diversity (does the generated data fit the whole data), and mode-collapse. We also evaluate the fidelity of the synthetic data using a transformers-based post-hoc classifier metric, Long-Sequence Discriminative Score (LDS). In addition, we use the Long-Sequence Predictive Score (LPS) to observe the predictive properties of synthetic data. We conduct another sequence prediction task to assess the predictive characteristics of the synthetic data. The goal of the task is to predict the next five steps of feature vectors given an input sequence.

We evaluate the quality of the synthetic data in terms of fidelity (does the generated data comes from the same distributions as the original one), diversity (is the generated data diverse enough to fit the whole data) and check if the generative model is demonstrating the mode-collapse problem or not. We also implement a transformers-based post-hoc classifier metric, Long-Sequence Discriminative Score (LDS) that evaluates the fidelity of the synthetic data. Additionally, we implement a transformers-based sequence prediction metric, Long-Sequence Predictive Score (LPS) to observe the predictive characteristics in the synthetic data. In order to evaluate the predictive characteristics of the synthetic data beyond LPS, we conduct another sequence prediction task. The goal of the task is to predict the next five steps of feature vectors given an input sequence.

\subsubsection{PCA \& t-SNE plots}

To evaluate the quality of the synthetic data, we use both Principal Component Analysis \cite{bryant1995principal}, and t-distributed Stochastic Neighbor Embedding \cite{van2008visualizing} techniques. They show how closely the distribution of synthetic data resembles real data in 2-dimensional space and also show if the generated data covers the area of the original data. This gives us the fidelity and diversity evaluation of the synthetic data.

%\begin{figure}
%	\centering
%	\includegraphics[width = 0.35\textwidth]{transfusion-output-gtgan-cotgan.png}
%	\caption{Sample of Synthetic data from GT-GAN, CoTGAN and TransFusion (ours) and original data on the Energy Dataset (dimension = 28), Sequence Length: 384}
%	\label{fig:result}
%\end{figure}

\subsubsection{Long-Sequence Discriminative Score}

We create an evaluation metric to determine if the synthetic data is indistinguishable from the original data. In order to use this metric, first we train a transformers-based \cite{vaswani2017attention} (only using a transformers encoder) post-hoc classifiers with the original \& synthetic data labeled as original and fake, respectively. Then we use the held-out dataset to measure the discriminative score by $discriminatve = \lvert 0.5 - accuracy \rvert$. Lower LDS means original and synthetic data has higher similarity. This metric is inspired by the discriminative score from the TimeGAN \cite{yoon2019time}. But the discriminative score from TimeGAN uses a RNN-based classifier, which do not work when the sequence length is long. Consequently, we use a transformer-based classifier to capture long-term time dependencies. This metric provides an evaluation of the fidelity of the assessment.

\begin{table*}[!h]
	\caption{Resutls on Different Time-Series Datasets, Sequence Length: 384, we report the mean and standard deviation (bold indicates best result, $\uparrow$ indicates higher as better, $\downarrow$ indicates lower as better)}
	\centering
	%	\rotatebox{90}{
		\resizebox{1.0\columnwidth}{!}{%
			\begin{tabular}{lcccccccc}\toprule
				& \multicolumn{7}{c}{\textbf{Evaluation Metrics}} 
				\\\cmidrule(lr){3-9}
				\textbf{Dataset} & \textbf{Model Name} & LDS & LPS & JSD & $\alpha$- & $\beta$- & Coverage & + 5 Steps\\ 
				& &  & & & precision & recall & & Ahead\\ 
				& & ($\downarrow$) & ($\downarrow$) & ($\downarrow$) & ($\uparrow$)& ($\uparrow$)& ($\uparrow$) & ($\downarrow$)\\ \midrule
				& GT-GAN  & 0.497 $\pm$ .003 & 0.313 $\pm$ .001& 0.189 $\pm$ .00 & 0.978 $\pm$ .00 & 0.016 $\pm$ .00 & 0.162 $\pm$ .00 & 0.316 $\pm$ .001\\
				Sine & CotGAN  & 0.370 $\pm$ .075 & 0.307 $\pm$ .000& 0.465 $\pm$ .00 & 0.904 $\pm$ .00 & 0.984 $\pm$ .00 & 0.926 $\pm$ .00 & 0.310 $\pm$ .001 \\
				& Sig-WGAN & 0.493 $\pm$ .001 & 0.311 $\pm$ .001& 0.551 $\pm$ .00 & 0.863 $\pm$ .001 & 0.567 $\pm$ .00 & 0.873 $\pm$ .00 & 0.312 $\pm$ .001 \\
				& TTS-GAN  & 0.5 $\pm$ .00 & 0.45 $\pm$ .001 & 0.542 $\pm$ .00 &  0.00 $\pm$ .00 & 0.00 $\pm$ .00 &  0.00 $\pm$ .00 & 0.443 $\pm$ .001\\
				& TimeGrad  & 0.498  $\pm$ .00 & 0.47 $\pm$ .001 &  0.59 $\pm$ .00 & 0.82 $\pm$ .00 &  0.00 $\pm$ .00 & 0.00 $\pm$ .00 &  0.48 $\pm$ .001\\
				& TransFusion (ours)  & \textbf{0.075} $\pm$ .032 & \textbf{0.306} $\pm$ .001 & \textbf{0.006} $\pm$ .00 & \textbf{0.99} $\pm$ .00 &  \textbf{0.99} $\pm$ .00 & \textbf{0.96} $\pm$ .00 &\textbf{ 0.301} $\pm$ .001\\
				\midrule
				& GT-GAN  &\textbf{ 0.441} $\pm$ .022 & 0.112 $\pm$ .000& 0.091 $\pm$ .00 & 0.961 $\pm$ .00 & 0.913 $\pm$ .00 & 0.556 $\pm$ .00 & 0.112 $\pm$ .001\\
				& CotGAN  & 0.445 $\pm$ .148 & 0.087 $\pm$ .000& 0.804 $\pm$ .00 & 0.840 $\pm$ .00 & 0.972 $\pm$ .00 & 0.870 $\pm$ .00 & 0.087 $\pm$ .001\\
				Stock& Sig-WGAN & 0.50 $\pm$ .00 & 0.151 $\pm$ .001& 0.77 $\pm$ .00 & 0.227  $\pm$ .00 & \textbf{0.998} $\pm$ .00 & 0.001 $\pm$ .00 & 0.149 $\pm$ .001 \\
				& TTS-GAN  &  0.5 $\pm$ .00 & 0.858 $\pm$ .001 & 0.781 $\pm$ .00 &  0.00 $\pm$ .00 & 0.00 $\pm$ .00 &  0.00 $\pm$ .00 & 0.858 $\pm$ .001\\
				& TimeGrad  &  0.5 $\pm$ .00 & 0.849 $\pm$ .001 & 0.08 $\pm$ .00 & 0.00 $\pm$ .00 &0.00  $\pm$ .00 &  0.00$\pm$ .00 &0.845  $\pm$ .001\\
				& TransFusion (ours)  & 0.483 $\pm$ .001 &\textbf{0.060}$\pm$ .001 & \textbf{0.018} $\pm$ .00 & \textbf{0.984} $\pm$ .00 &  0.991 $\pm$ .00  &\textbf{ 0.984} $\pm$ .00 &\textbf{ 0.061} $\pm$ .001\\
				\midrule
				
				& GT-GAN  & 0.495 $\pm$ .004 & 0.002 $\pm$ .000& 0.182 $\pm$ .00 & \textbf{1.00} $\pm$ .00 & 0.888 $\pm$ .00 & 0.190 $\pm$ .00 & 0.003 $\pm$ .001\\
				& CotGAN  & 0.481 $\pm$ .030 & 0.006 $\pm$ .000& 0.780 $\pm$ .00 & 0.980 $\pm$ .00 & 0.304 $\pm$ .00 & 0.920 $\pm$ .00 & 0.005 $\pm$ .001\\
				Air & Sig-WGAN & 0.5 $\pm$ .00 & 0.064 $\pm$ .001& 0.510 $\pm$ .00 & 0.00 $\pm$ .00 & 0.329 $\pm$ .00 & 0.00 $\pm$ .00 & 0.064 $\pm$ .001 \\
				& TTS-GAN  &  0.5 $\pm$ .00 & 0.934 $\pm$ .001 &0.436  $\pm$ .00 &  0.00$\pm$ .00 & 0.00$\pm$ .00 & 0.00  $\pm$ .00 &  0.934 $\pm$ .001\\
				& TimeGrad  & 0.5 $\pm$ .00 & 0.535 $\pm$ .223 & 0.831 $\pm$ .00 & 0.00 $\pm$ .00 & 0.504 $\pm$ .00 & 0.00 $\pm$ .00 & 0.44 $\pm$ .23\\
				& TransFusion (ours)  &\textbf{ 0.162} $\pm$ .029 & \textbf{0.001} $\pm$ .001& \textbf{0.027} $\pm$ .00 & 0.951 $\pm$ .00 & \textbf{0.956} $\pm$ .00 & \textbf{0.944} $\pm$ .00 & \textbf{0.002} $\pm$ .001\\
				
				\midrule
				
				& GT-GAN  & 0.434 $\pm$ .041 & 0.012 $\pm$ .000& 0.074 $\pm$ .00 & 0.586 $\pm$ .00 & 0.334 $\pm$ .00 & 0.480 $\pm$ .00 & 0.012 $\pm$ .001\\
				& CotGAN  & 0.472 $\pm$ .024 & 0.012 $\pm$ .000 & 0.529 $\pm$ .00 & \textbf{1.000} $\pm$ .00 & 0.00 $\pm$ .00 & 0.096 $\pm$ .00 & 0.012 $\pm$ .001\\
				Energy& Sig-WGAN & 0.5 $\pm$ .00 & 0.028 $\pm$ .001& 0.79 $\pm$ .00 & 0.00 $\pm$ .00 & 0.00 $\pm$ .00 & 0.00 $\pm$ .00 & 0.03 $\pm$ .001 \\
				& TTS-GAN  &  0.5 $\pm$ .00 & 0.81 $\pm$ .189 & 0.802 $\pm$ .00 & 0.00 $\pm$ .00 & 0.00 $\pm$ .00 &  0.00 $\pm$ .00 & 0.728 $\pm$ .275\\
				& TimeGrad  &  0.5 $\pm$ .00 & 0.970 $\pm$ .001 &  0.14 $\pm$ .00 & 0.00 $\pm$ .00 & 0.00 $\pm$ .00 &  0.00$\pm$ .00 & 0.970 $\pm$ .00\\
				& TransFusion (ours)  & \textbf{0.400} $\pm$ .019 & \textbf{0.011} $\pm$ .001 & \textbf{0.012} $\pm$ .00 & 0.998 $\pm$ .00 & \textbf{0.644} $\pm$ .00 & \textbf{0.850} $\pm$ .00 & \textbf{0.011}$\pm$ .001\\
				
				\bottomrule
			\end{tabular}
		}
		\label{tab:evaluation-table2}
		%}
\end{table*}

\subsubsection{Long-Sequence Predictive Score}

To check if the synthetic data captured the predictive characteristics of the original data or not, we create a transformers-based sequence prediction task called long-sequence predictive score (LPS). Here we train a transformers-based sequence prediction model to predict the next step feature vector given an input sequence. The metric was also inspired by TimeGAN's predictive score \cite{yoon2019time}, but since TimeGAN's predictive score uses a RNN-based model, it is not be suitable for long-sequenced tasks. We train the model using synthetic data and then evaluate it using the original dataset. The Mean Absolute Error (MAE) is used to measure the performance of the LPS. In addition, we extend the LPS to predict the next five steps (\textbf{+5 Steps Ahead}) feature vectors given the input sequence to see how features evolve over time.

\subsubsection{Jensen-Shannon Divergence}

This quantitative evaluation metric measures if the original and the synthetic data come from the same distributions or not. A lower JSD \cite{menendez1997jensen} score means the distributions are similar between the original and synthetic datasets.

\subsubsection{$\alpha$-precision, $\beta$-recall \& Coverage}

$\alpha$-precision and $\beta$-recall \cite{alaa2022faithful} also provide the assessment of fidelity and diversity, respectively, of the synthetic data. And coverage \cite{alaa2022faithful} checks the mode-dropping issue in the generative models by using nearest neighbor manifolds and measuring the fraction of the original samples whose neighborhoods contain at least one synthetic sample. %If the generative models fail to learn the original data distribution then the generated samples will not align 

\subsection{Visual Evaluation}
\begin{figure}[!h]
	\centering
	\includegraphics[width = 0.8\textwidth]{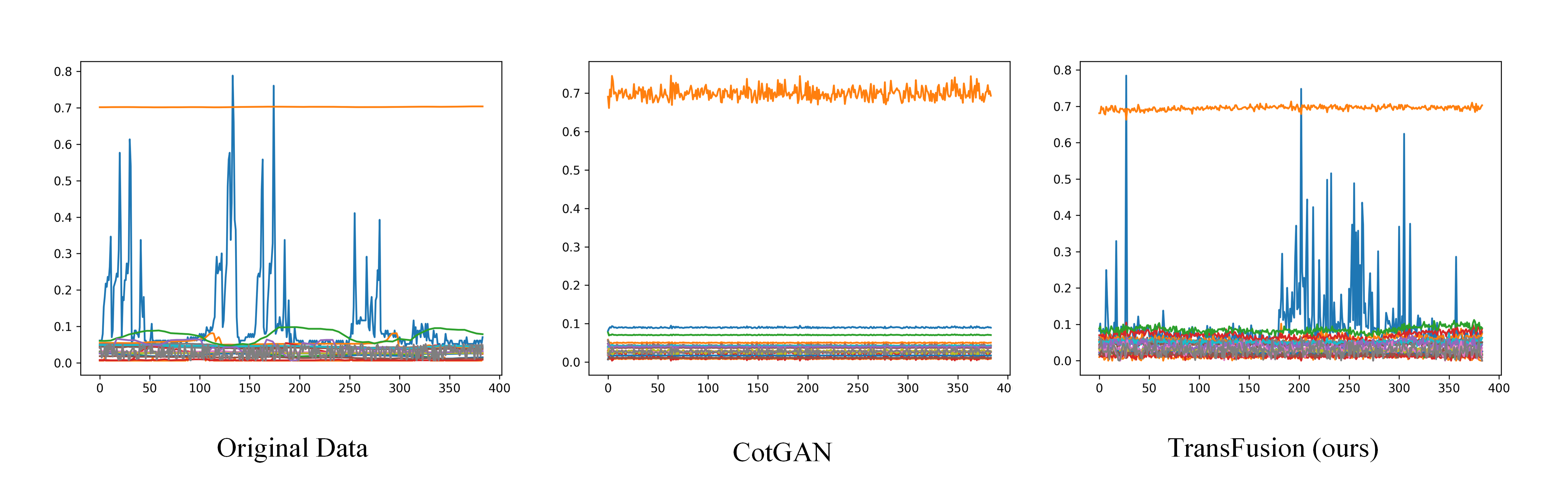}
	\caption{Comparison with original data and generated samples by CotGAN and TransFusion, Sequence length: 384, Energy dataset}
	\label{fig:fusion-output}
\end{figure}

\begin{figure*}[t]
	\centering
	\includegraphics[width = 0.8\textwidth]{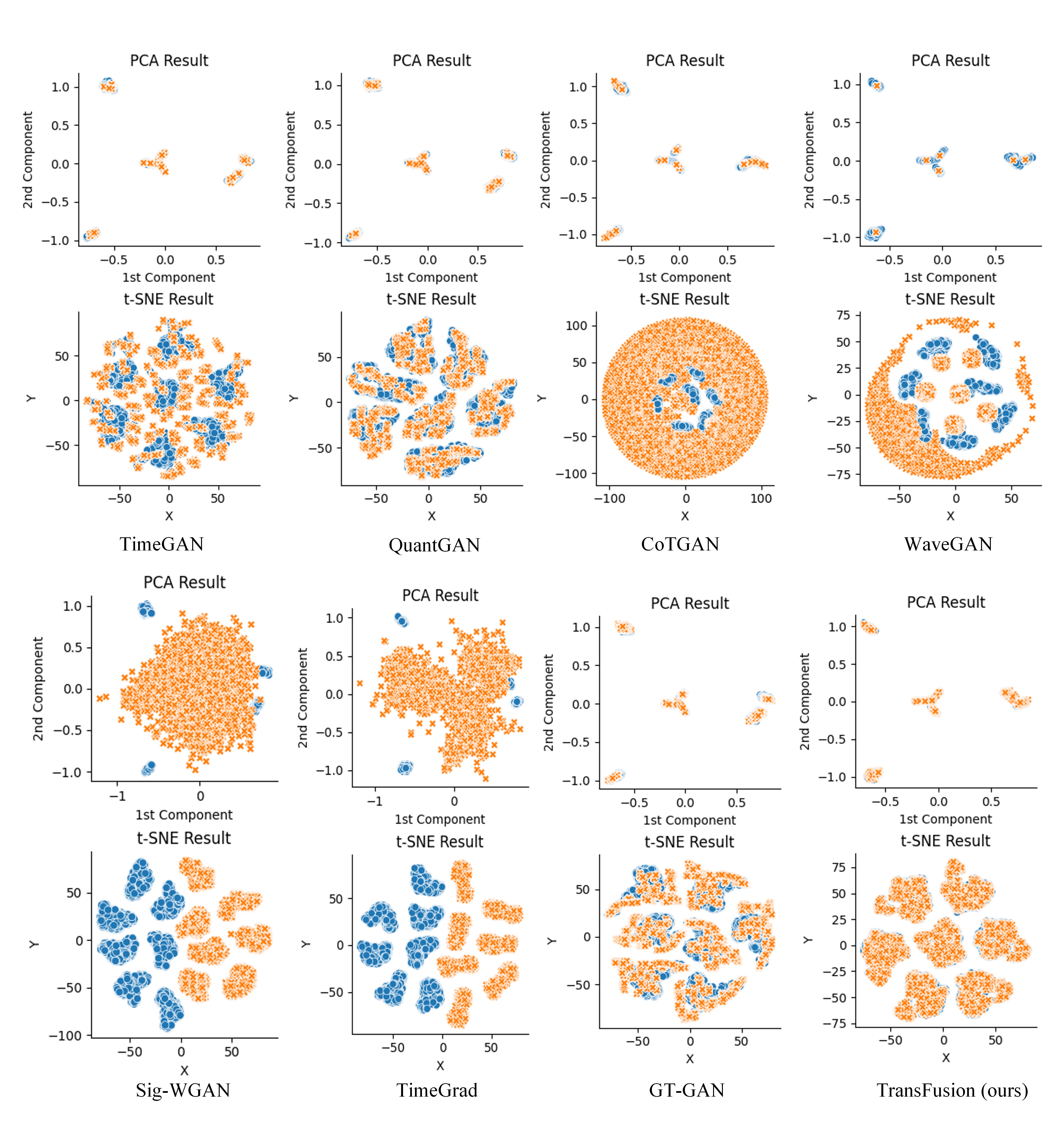}
	\caption{PCA and t-SNE plots of TimeGAN, QuantGAN, CoTGAN, WaveGAN, Sig-WGAN, TimeGrad, GT-GAN, and TransFusion (ours) on the Energy Dataset (dimension = 28), Sequence Length: 100. Each dot represents a sequence of time series data, blue and orange represents real and synthetic data respectively, if the generative model learns to approximate the real data distributions, the blue and orange dots should overlap}
	\label{fig:pca-tsne-100}
\end{figure*}

For the qualitative comparison, we use the PCA and t-SNE plots of our generative models and compare them with other model plots. Figure \ref{fig:pca-tsne-100}, shows the plots of the energy dataset, sequence length 100 (high-dimensional data in our experiment). Each dot in the plot represents a sequence of data. We can observe, for TransFusion, the generated data's dot is closer to the original data. This indicates that our generative model learns the distribution of the original data and is capable of generating high-quality data. Figure \ref{fig:fusion-output} compares the original data and generated samples by CotGAN and Transfusion. We can see that TransFusion can generate great results even with a higher sequence length (384), whereas the CoTGAN cannot keep up.

\subsection{Empirical Evaluation}

An ideal generative model should generate samples that are statistically close to the original data (fidelity) and diverse enough to cover the whole area of the original data (diversity), and the model should not fall into mode dropping (coverage). We provide multiple assessments of fidelity (LDS, JSD, $\alpha$-precision, PCA plots, t-SNE plots) and diversity ($\beta$-recall, PCA plots, t-SNE plots) and also show if the generative models are suffering from the mode-collapse problem or not (coverage). We also show the predictive characteristicness of the synthetic data by using LPS and +5 Steps Ahead score. 

In our experiments, we use both high dimensional and long-sequenced data (Air Quality, Energy Consumption). Table \ref{tab:evaluation-table1} indicates the results of our experiment for sequence length 100. We run the evaluation metrics ten times and report the mean and standard deviation in the table. Overall, TransFusion performs better than all the state-of-the-art generative models. We achieve a significantly higher score in every metric. Although some of the other methods get a high score in some metrics, but if we consider all the aspects of fidelity, diversity, and coverage, we observe that TransFusion outperforms them all. The combination of diffsuion model and Transformers make it possible to generate high-dimensional and long-sequenced data. Whereas, Generative Adversarial Networks based architecture provide limited sample variety due to the mode-collapse problem, also as they use CNN, RNN-based architecture, they fail to generate long-sequenced data. 

\begin{table*}[!htbp]
	\caption{LDS vs Discriminative Score \cite{yoon2019time} performance on TransFusion: Sequence Length: 384, we report the mean and standard deviation of both LDS and Discriminative score, also we report the F1 score of each class (0 and 1)}
	\centering
	\resizebox{0.7\columnwidth}{!}{%
		\begin{tabular}{lcccccc}\toprule
			& \multicolumn{6}{c}{\textbf{Evaluation Metrics}}
			\\\cmidrule(lr){2-7}
			& \multicolumn{3}{c}{\textbf{LDS}} & \multicolumn{3}{c}{\textbf{Discriminative Score}}
			\\\cmidrule(lr){2-4} \cmidrule(lr){5-7}
			\textbf{Dataset} & \ \textbf{Score} & \multicolumn{2}{c}{\textbf{F1 Score}} & \textbf{Score} & \multicolumn{2}{c}{\textbf{F1 Score}} \\
			\cmidrule(lr){3-4} \cmidrule(lr){6-7}
			& & 0 & 1 &  &0 & 1\\\midrule
			Sine &  0.075 $\pm$ .032 & 0.985 $\pm$ .00 & 0.984 $\pm$ .00 & 0.032 $\pm$ .019 & 0.528 $\pm$ .00 & 0.459 $\pm$ .00 \\ \midrule
			Stock &  0.483 $\pm$ .001  & 1.00 $\pm$ .00 & 1.00 $\pm$ .00 & 0.276 $\pm$ .191 & 0.67 $\pm$ .00 & 0.00 $\pm$ .00 \\ \midrule
			Air &  0.162 $\pm$ .029 & 0.72 $\pm$ .00 & 0.731 $\pm$ .00 & 0.158 $\pm$ .003 & 0.652 $\pm$ .00 & 0.468 $\pm$ .00 \\ \midrule
			Energy &  0.400 $\pm$ .019 & 0.978 $\pm$ .00 & 0.975 $\pm$ .00 & 0.215 $\pm$ .115 & 0.692 $\pm$ .00 & 0.256 $\pm$ .00 \\ \midrule 
			
		\end{tabular}
		\label{tab:ldsvsdisc}
	}
\end{table*}

\begin{table*}[!htbp]
	\caption{Ablation Study: Sequence Length: 100, we report the mean and standard deviation (bold indicates best result, $\uparrow$ indicates higher as better, $\downarrow$ indicates lower as better)}
	\centering
	%	\rotatebox{90}{
		\resizebox{1.0\columnwidth}{!}{%
			\begin{tabular}{lcccccccc}\toprule
				& \multicolumn{7}{c}{\textbf{Evaluation Metrics}} 
				\\\cmidrule(lr){3-9}
				\textbf{Dataset} & \textbf{Model Name} & LDS & LPS & JSD & $\alpha$- & $\beta$- & Coverage & + 5 Steps\\ 
				& &  & & & precision & recall & & Ahead\\ 
				& & ($\downarrow$) & ($\downarrow$) & ($\downarrow$) & ($\uparrow$)& ($\uparrow$)& ($\uparrow$) & ($\downarrow$)\\ \midrule
				
				& TransGAN  & 0.500 $\pm$ .000 & 0.352 $\pm$ .000& 0.392 $\pm$ .00 & 0.000 $\pm$ .00 & 0.000 $\pm$ .00 & 0.000 $\pm$ .00 & 0.349 $\pm$ .001\\
				Sine & Diffusion-GRU & 0.498 $\pm$ .002 & 0.252 $\pm$ .000& 0.106 $\pm$ .00 & 0.946 $\pm$ .00 & 0.408 $\pm$ .00 & 0.703 $\pm$ .00 & 0.260 $\pm$ .001\\
				& TransFusion (ours)  & \textbf{0.097} $\pm$ .042 & \textbf{0.243} $\pm$ .001 & \textbf{0.02} $\pm$ .00 & \textbf{0.992} $\pm$ .00 &  \textbf{0.974} $\pm$ .00 & \textbf{0.963} $\pm$ .00 & \textbf{0.245} $\pm$ .001\\ \midrule
				
				& TransGAN  & 0.500 $\pm$ .000 & 0.905 $\pm$ .011& 0.630 $\pm$ .00 & 0.000 $\pm$ .00 & 0.000 $\pm$ .00 & 0.000 $\pm$ .00 & 0.862  $\pm$ .001\\
				Stock & Diffusion-GRU  & 0.500 $\pm$ .000 & 0.066 $\pm$ .000& 0.273 $\pm$ .00 & 0.315 $\pm$ .00 & 0.744 $\pm$ .00 & 0.166 $\pm$ .00 & 0.067 $\pm$ .001\\
				& TransFusion (ours)  & \textbf{0.498} $\pm$ .001 &\textbf{0.060}$\pm$ .001 & \textbf{0.018} $\pm$ .00 & \textbf{0.97} $\pm$ .00 &  \textbf{0.99} $\pm$ .00  &\textbf{ 0.93} $\pm$ .00 & \textbf{0.061} $\pm$ .001\\ \midrule
				
				& TransGAN  & 0.500 $\pm$ .000 & 0.065 $\pm$ .001& 0.532 $\pm$ .00 & 0.000 $\pm$ .00 & 0.000 $\pm$ .00 & 0.000 $\pm$ .00 & 0.080 $\pm$.001\\
				Air & Diffusion-GRU  & 0.500 $\pm$ .000 & 0.051 $\pm$ .005& 0.230 $\pm$ .00 & 0.001 $\pm$ .00 & 0.049 $\pm$ .00 & 0.001 $\pm$ .00 &  0.057 $\pm$ .001 \\
				& TransFusion (ours)  &\textbf{ 0.148} $\pm$ .029 &\textbf{ 0.001} $\pm$ .001& \textbf{0.027} $\pm$ .00 & \textbf{0.926 }$\pm$ .00 & \textbf{0.956} $\pm$ .00 & \textbf{0.944} $\pm$ .00 & \textbf{0.002} $\pm$ .001\\ \midrule
				
				& TransGAN  & 0.500 $\pm$ .000 & 0.971 $\pm$ .000& 0.692 $\pm$ .00 & 0.000 $\pm$ .00 & 0.000 $\pm$ .00 & 0.000 $\pm$ .00 & 0.970 $\pm$ .001\\
				Energy & Diffusion-GRU  & 0.500 $\pm$ .000 & 0.097 $\pm$ .007& 0.421 $\pm$ .00 & 0.000 $\pm$ .00 & 0.000 $\pm$ .00 & 0.000 $\pm$ .00 & 0.092 $\pm$ .001 \\
				& TransFusion (ours)  & \textbf{0.332} $\pm$ .019 & \textbf{0.011} $\pm$ .001 & \textbf{0.016} $\pm$ .00 & \textbf{0.983} $\pm$ .00 & \textbf{0.876} $\pm$ .00 & \textbf{0.847} $\pm$ .00 & \textbf{0.012}$\pm$ .001\\
				\bottomrule
			\end{tabular}
		}
		
		\label{tab:ablation-table1}
		%}
\end{table*}

From Table \ref{tab:evaluation-table1}, we can observe our evaluation metrics, LDS and LPS reflect the results with other metrics. Transfusion achieve a 22\% higher score than the GT-GAN on the Long-Sequence Discriminative Score for the energy dataset. Also, from the coverage metric, we see that, except TransFusion, all the models suffer mode-collapse when the data is high dimensional (energy dataset). So, these state-of-the-art methods can not keep up with the long-sequence high-dimensional data. On top of these, the synthetic data generated by TransFusion achieve the best score on both LPS and +5 steps ahead prediction tasks. Therefore, synthetic data generated by TransFusion can be used in the forecasting tasks.

Furthermore, we also stretch the sequence length of the datasets to 384 to observe the robustness of our model and compare the result with GT-GAN, CotGAN, Sig-WGAN, TTS-GAN and TimeGrad. Table \ref{tab:evaluation-table2}, shows the outcome. We see that, all of these models can not keep up with the long-sequence data, and suffer the mode-collapse problem. On top of that, TimeGrad, though being a diffusion-based model, cannot generate high-fidelity, long-sequenced data due to the usage of RNN-based architecture.

%We also show the generated samples by TransFusion and comapre this with GT-GAN and CoTGAN's generated data. Figure \ref{fig:result} shows the comparison. We see, the generated samples by TransFusion is close to the original data and other models cannot generate high-quality samples when the sequence length and dimension is high. 

Besides empirical evaluation, training time is another factor to consider. TransFusion is able to generate high-quality samples by only training for 5 hours whereas GT-GAN takes nearly 28 hours to produce their result. All the experiments were run on an NVIDIA A100 GPU with 40GB of memory.

%\begin{table*}[!h]
%	\caption{Resutls on Different Time-Series Datasets, Sequence Length: 384, we report the mean and standard deviation (bold indicates best result, $\uparrow$ indicates higher as better, $\downarrow$ indicates lower as better)}
%	\centering
%	%	\rotatebox{90}{
	%		\resizebox{0.5\columnwidth}{!}{%
		%			\begin{tabular}{lccccccc}\toprule
			%				& \multicolumn{3}{c}{\textbf{Evaluation Metrics}} 
			%				\\\cmidrule(lr){3-8}
			%				\textbf{Dataset} & \textbf{Model Name} & LDS & Discriminative Score\\ 
			%				%& &  & & & precision & recall & & Ahead\\ 
			%				%& & ($\downarrow$) & ($\downarrow$) & ($\downarrow$) & ($\uparrow$)& ($\uparrow$)& ($\uparrow$) & ($\downarrow$)\\ \midrule
			%
			%				
			%				\bottomrule
			%			\end{tabular}
		%		}
	%		%	
	%		\label{tab:lds-discriminative}
	%		%}
%\end{table*}

\subsection{LDS vs Discriminative Score}

To show the effectiveness of Long-Sequence Discriminative score against the the Discriminative score \cite{yoon2019time}, we conduct an experiment. As the both of these evaluation metrics is basically a classification method so we calculate the F1-score to see how are they performing for individual class. We take the generated data by TransFusion (sequence length 384) and calculate the F1 score for both LDS and Discriminative score. From Table \ref{tab:ldsvsdisc}, we can see, though for some dataset the Discriminative Score is performing better (lower the better score) but F1-score of individual class is poor comparison with LDS. Discriminative score could not classify properly when the sequence length was longer due to its RNN architecture.

\subsection{Ablation Study}

The combination of the diffusion model and transformers architecture allows us to generate long-sequenced, high-quality samples, as evidenced by the ablation study. For this, we train a transformers-based Generative Adversarial Network (TransGAN), where the transformer's encoder serves as both generator and discriminator. We also train a diffusion model without transformers. For the diffusion model's backward process, we use the Gated Recurrent Unit (GRU). Table \ref{tab:ablation-table1} shows the analysis of the experiment. We use sequence length 100 for the ablation study. We can see both TransGAN and Diffusion-GRU is unable to keep up with the TransFusion. The synthetic data generated by TransGAN and Diffusion-GRU do not correlate with the original data distributions because these two models failed to capture the data distribution. As a result, for most of the evaluation metrics, they score poorly due to absence of \textit{true positive} data in the synthetic data \cite{alaa2022faithful}.

\section{Limitations and Future Work}

While diffusion-based generative models are fast in terms of training compared with other generative models such as GANs, sampling from learned distributions takes longer. But, on the other hand, it learns the data distribution well and is capable of generating high-quality data, and also overcomes the mode-collapse problem. Variational AutoEncoder (VAE) can sample faster and avoid mode-collapse problem but is incapable of generating high-quality samples. GANs can generate high-quality samples faster but are prone to mode-collapse.

In future work, beside generating high-quality, long-sequenced time-series, we will try to generate fair synthetic time-series data.

\section{Conclusion}
\label{sec:conclusion}

In this study, we present a diffusion and transformers-based framework, \textbf{TransFusion}, for generating high-quality, long-sequence time-series data. Generating long-sequence time-series data is important because of its numerous applications. Long sequence data can capture more context and pattern of the data better than shorter sequences. In the past, researchers used, Generative Adversarial Network-based generative models to generate time-series data, but GAN is well known for falling into mode-collapse problems, and usage of standalone Recurrent Neural Network- and Convolutional Neural Network-based architectures made it difficult to generate long-sequence data. Our framework, with the combination of a transformers architecture and diffusion model, solve both mode collapse problems and capture longer time dependencies. We also implement two new evaluation metrics for time-series synthetic data. We evaluate our framework with various evaluation metrics and show it outperforms state-of-the-art architectures. TransFusion successfully generated high-fidelity data with a sequence length of 384. We believe this framework can generate even longer time-series data given adequate GPU memory. This framework can be used in different domains to generate synthetic time-series data to better understand the data context and find patterns in the data.

\section*{Acknowledgments}

This work was funded by the Knut and Alice Wallenberg Foundation, the ELLIIT Excellence Center at Linköping-Lund for Information Technology (portions of this work were carried out using the AIOps/Stellar), and TAILOR - an EU project with the aim to provide the scientific foundations for Trustworthy AI in Europe. The computations were enabled by the Berzelius resource provided by the Knut and Alice Wallenberg Foundation at the National Supercomputer Centre.

%\section*{Acknowledgments}
%This was was supported in part by......

%Bibliography
\bibliographystyle{unsrt}  
\bibliography{templateArxiv}

\end{document}